\documentclass[11pt]{article}

\usepackage[final]{acl}

\usepackage{times}
\usepackage{latexsym}
\usepackage{booktabs} 

\usepackage[T1]{fontenc}

\usepackage[utf8]{inputenc}

\usepackage{microtype}

\usepackage{inconsolata}

\usepackage{graphicx}

\usepackage{listings}
\title{lmfaoooo at SemEval-2026 Task 1: Humor Is an Audience. \\ Preference Modeling for Constrained Humor Generation}

\author{Alexey Tikhonov \\
  Inworld.AI \\
  Berlin, Germany \\
  \texttt{altsoph@gmail.com} \\\And
  Alexey Ivanov \\
  OpenAI \\
  Mountain View, CA \\
  \texttt{SaveTheRbtz@GMail.com} \\}

\begin{document}
\maketitle
\begin{abstract}
Humor generation remains difficult not only because producing fluent, novel jokes is hard, but because ``funny'' is audience-dependent and supervision is noisy—preferences vary with audience, context, and culture, and annotator agreement is often low.
In this paper, we describe our system for the SemEval-2026 Task~1 (MWAHAHA) \cite{semeval2026mwahaha}, which focuses on humor generation under explicit constraints. The task evaluates submitted systems via human preference judgments in 1-on-1 arena-style comparisons.

We adopt a ``generate-many $\rightarrow$ select-best'' strategy. First, we generate a diverse pool of candidates per instance using multi-step prompting, model ensembling, and diversity-oriented decoding. Second, we select outputs using a preference model that approximates a ``reader'' by learning from human comparisons rather than absolute funniness scores. To support this approach, we release $\sim$2.5K human pairwise judgments collected through the Humor Arena prototype \cite{Ivanov2025HumorArena}. We further propose an interpretable pipeline that converts labeled comparisons into a preference model. Across three preference datasets, our models consistently outperform baselines and show stronger cross-domain transfer. Finally, we apply the learned preference model to rank candidates for the MWAHAHA setting and release intermediate artifacts (candidate pools and rankings) to facilitate follow-up work. \footnote{\url{https://github.com/altsoph/lmfaoooo}} 

Our system ranked 1st in the English and Chinese subtasks of MWAHAHA and 2nd in the Spanish subtask.
\\ \\ \\ \\

\end{abstract}

\section{Introduction}
\begin{quote}\small
\emph{``The question is,'' said Alice, ``whether you can make words mean so many different things.''\\
``The question is,'' said Humpty Dumpty, \\ ``which is to be master---that’s all.''} \cite{Carroll1871LookingGlass}
\end{quote}

Can computers be funny? 
Humor generation has long been viewed as a stress test for natural language generation: beyond grammatical and semantic coherence, a successful joke must trigger an audience response that is strongly shaped by culture, demographics, and context. Different authors, from \cite{Propp1976ProblemyKomizmaISmekha} to \cite{warren2021}, describe many competing theories  of humor appreciation and emphasize that humor depends on incongruity, violation/benign framing, and the audience. A practical consequence is that humor supervision is inherently noisy: even careful human labeling setups often show limited agreement, and different groups systematically disagree about what they prefer \cite{Ruch2013Assessment3WD}, \cite{murakami2026laughswhomdisentanglinginfluential}. 
Even in controlled dataset settings, annotators disagree substantially; moderate agreement is common in humor-related annotation tasks, e.g., Fleiss' $\kappa \approx 0.49$ for pun-related data \cite{sun-etal-2022-expunations}. As a result, ``gold labels'' are often better viewed as samples from a preference distribution than as objective truth.


A second practical challenge is memorization and mode collapse in joke generation. In a systematic probe of ChatGPT’s joke outputs \cite{jentzsch23}, over 90\% of 1008 generated jokes repeated the same 25 jokes, suggesting that unconstrained generation can default to a small set of high-frequency templates rather than produce novel humor. This motivates evaluation and selection strategies that reward diversity and penalize near-duplicates. Moreover, models are often compared against “polished” jokes that have been socially selected (e.g., through comedian curation or online popularity), whereas AI outputs are largely unfiltered—an asymmetry that can bias evaluations against the model.

SemEval-2026 Task~1, MWAHAHA (Models Write Automatic Humor And Humans Annotate) \cite{semeval2026mwahaha}, is the first SemEval task dedicated to advancing computational humor generation, including text-based constrained joke generation. The task aims to push models beyond memorization by requiring generation under constraints and evaluating outputs with human annotation. In this work, we argue that for constrained humor generation, the bottleneck is often selection rather than raw generation: modern LLMs can produce many plausible candidates, but reliably identifying which candidate will be funniest for the target audience is the core difficulty. We therefore treat humor generation as a two-stage process: (i) generate diverse candidates, then (ii) select the optimal one using learned audience preferences.

Our main contributions are:
\begin{itemize}
  \item \textbf{Preference data release.} We release $\sim$2.5K human pairwise judgments from Humor Arena \cite{Ivanov2025HumorArena}.
  \item \textbf{Preference-first framing.} We formalize constrained humor generation as generation and ranking of multiple candidates using repeated pairwise preference decisions and show this yields more stable learning signals than baseline scoring.
  \item \textbf{Interpretable preference model.} We propose an interpretable ``humor feature'' extraction pipeline that converts comparisons into a compact feature basis and enables training of lightweight preference models.
  \item \textbf{lmfaoooo system.} We apply the pipeline to SemEval-2026 Task~1 and document our full generation and selection workflow. We publicly release\footnote{\url{https://github.com/altsoph/lmfaoooo}}  our prompts, candidate pools, features, and rankings used in our experiments, as long as versions of the used models and other information to support reproducibility.
\end{itemize}

\section{Background: humor, novelty, and evaluation}
Humor is not a single phenomenon but a family of mechanisms (e.g., incongruity, wordplay, expectation violation). Surveys of humor theories enumerate many partially overlapping accounts, suggesting that a universal definition of ``funniness'' is unlikely to be operationally sufficient for model training and evaluation \cite{warren2021}. Classic humor studies also report differences in humor appreciation across demographic groups, reinforcing that evaluation depends on the rater population \cite{MundorfEtAl1988GenderDifferencesHumor}, \cite{Dore2019HumourAVT}.

\cite{TikhonovS24} show, in blind evaluations, that careful task formulation can elicit machine-generated humor of a quality comparable to human jokes, including through the use of brainstorming-style prompting techniques. A common pattern in creative production is to generate many candidates and then select the best. In LLM creativity studies, explicitly separating brainstorming from selection improves performance over brainstorming alone \cite{SummersStay2023BrainstormTS}. Different tricks can be used to increase the diversity of generations \cite{zhang2025verbalizedsamplingmitigatemode}. Moreover, direct training for joke generation also yields strong results \cite{WangEtAl2025InnovativeThinkingInfiniteHumor}. 

Effective and unbiased evaluation is a tricky problem in the era of massive adoption of LLMs \cite{post}. However, some interesting results are already achieved even for a subjective task of creative text evaluation \cite{Agafonova}. Automatic humor recognition, detection, or selection is also a known open problem \cite{Rada}, with some recent promising results \cite{Kalloniatis2024ComputationalHumorRecognitionSLR}, \cite{Bago2025LLMsHumorDetectionCroatian}. In subjective NLG tasks, pointwise (Likert-style) scoring is often noisy or inconsistent. Comparative protocols can be better aligned with human judgments in many settings \cite{novikova2018rankme}, although they introduce their own biases and costs. MWAHAHA \cite{semeval2026mwahaha} itself adopts a pairwise arena-style human preference evaluation for final scoring. Learning ranking from paired comparisons has a long tradition in statistics (e.g., Bradley--Terry style models \cite{BradleyTerry1952RankAnalysisPairedComparisons}). A practical advantage is that one can infer global rankings even with incomplete comparison graphs, which is useful when comparisons are expensive. Recent studies on pairwise LLM evaluation and rank aggregation (e.g., uncertainty-guided comparison scheduling and evaluator design) provide complementary evidence and actionable recommendations for improving both accuracy and cost.


This motivates our system design: treat candidate generation as a diversity-maximization problem and selection as a preference-learning problem; then, generate the final ranking using Evalica \cite{Ustalov:25}, an open-source toolkit designed to support reliable, reproducible evaluation and ranking.

\section{Task description: SemEval-2026 MWAHAHA} 
MWAHAHA Subtask~A \cite{semeval2026mwahaha} requires generating jokes under constraints (e.g., conditioning on a news headline or including specified words) in multiple languages (English, Spanish, Chinese). The task aims to encourage novelty and fairness through constraint design and human evaluation. 

\section{Data}
We train and evaluate preference models on three datasets.

\subsection{Reddit-based comparisons}
We use a Reddit jokes dataset with human ratings and comparisons derived from up/downvotes, originally collected by \cite{weller-seppi-2019-humor}. This dataset represents relatively broad, community-driven humor with strong topical and stylistic regularities. Since vote-derived labels can reflect confounds (e.g., visibility and temporal effects) rather than pure funniness, we treat them as noisy preference signals and take care to reduce near-duplicate leakage across train/test splits. The dataset was downloaded using scripts provided by \cite{baranov-etal-2023-told} and processed exactly the same way as described in \cite{TikhonovS24}.

\subsection{Humor Mechanics dataset}
We also use the publicly available human comparison data released with the Humor Mechanics project repository \footnote{\url{https://github.com/altsoph/humor-mechanics}}. It contains both human-written and generated jokes with human labels for them collected using assessors recruited via the Scale.AI service.

\subsection{Humor Arena pairwise judgments}
We collect pairwise judgments in Humor Arena\footnote{\url{https://humor.ph34r.me/}}, a lightweight platform for comparing jokes generated by different models and forming model/joke rankings via repeated comparisons. We release\footnote{\url{https://github.com/SaveTheRbtz/humor}} a dataset of 2{,}543 pairwise comparisons over generated one-liners collected in 391 sessions. Each comparison has one of four outcomes: A wins, B wins, both are good, or both are bad. For training preference models, we encode wins/losses as \(y\in\{1,0\}\) and discard both-good/both-bad outcomes.

\section{Method}

\subsection{Candidate generation}
For candidate generation stage, we closely follow a multi-step generation strategy described in \cite{TikhonovS24} and inspired by \cite{SummersStay2023BrainstormTS}: instead of attempting to produce the joke in one pass, we first generate, expand and refine a list of associations, then in the final phase we merge them into several candidate jokes (check Appendix B. of \cite{TikhonovS24} for the specific prompts).

To increase diversity and reduce memorization, we use (i) an ensemble of models (specifically claude-sonnet-4 \cite{Anthropic2025Claude4SystemCard}, claude-opus-4.5 \cite{Anthropic2025ClaudeOpus45SystemCard} and GPT-5 \cite{SinghEtAl2025OpenAIGPT5SystemCard}, all with the default temperature = 1.0), and (ii) a tail-focused sampling trick \cite{zhang2025verbalizedsamplingmitigatemode} designed to draw candidates from lower-probability regions of the model distribution. In our experiments, we generate 50 candidates per task instance (before deduplication and filtering). To filter the pool, we apply deterministic checks to enforce hard constraints of the MWAHAHA task (keyword inclusion and length limits) and discard invalid candidates. To reduce near-duplicate leakage and improve selection robustness, we additionally de-duplicate candidates using embedding similarity, keeping the centroid candidate per near-duplicate cluster.

\subsection{Building the ``humor basis''}
To select the best candidate, we need to compare them and get results as close as possible to the human audience. A core difficulty in preference modeling is representation: raw text embeddings can work, but they are hard to interpret and diagnose. We therefore construct a compact, interpretable feature ``humor basis'', a key element of our proposed approach -- a relatively short list of different humor aspects that can be used (or not) in a particular joke. For example, among such rules, there could be recommendations to use \emph{Dark Humor}, \emph{Exaggeration}, or \emph{Wordplay}. To some degree, it is similar to the Constitutional AI approach from \cite{const}, but we do not require the rules to be co-aligned or coherent.

At this stage, our goal is to achieve maximum diversity of popular humor aspects without overinflating the list. To construct such a basis, we extract qualitative ``difference hints'' from each training preference pair by prompting an LLM to produce 7--10 short descriptions of how the two jokes differ. This yields several thousand candidate hints. We then embed the hints, cluster them with DP-means \cite{KulisJordan2012DPMeans}, and discard small clusters. Finally, we apply semantic (LLM-based) de-duplication across clusters to obtain a compact basis (17 features in the current run; see Appendix A for the full list).

\subsection{Feature vector extraction}
Now, using the humor basis, we may use it for a feature vector extraction procedure by exposing a joke or pair of jokes aside with the basis and asking LLM to score it across the basis rules. We implement two different setups here: (i) pointwise: here we ask LLM to score a given joke independently by producing a vector of weights (one per rule), (ii) pairwise: here we ask LLM to compare two jokes and return as a vector of differences -- which joke in a pair is more aligned with the given rule. Check Appendix B and Appendix C for examples of pointwise and pairwise decompositions, respectively. Reach the publicly available github repo for exact decomposition prompts\footnote{\url{https://github.com/altsoph/lmfaoooo}}.

\subsection{Preference modeling}

Having these (pointwise and pairwise) feature vectors, we can now train lightweight preference predictors. We first use L1 (LASSO) regularization \cite{Tibshirani1996Lasso} for feature selection and collinearity reduction; then we train a L2-regularized regression \cite{Tikhonov1963RegularizationMethod} over selected features for better solution stability.

While pairwise feature vectors are more expensive (requiring up to $\Theta(n^2)$ evaluations to obtain a useful ranking), they often yield better self-consistency and stability \cite{novikova2018rankme} in global ranking.
Thus, we use a pairwise-based regression as a main approach, providing some optimizations to reduce the real number on comparisons: (i) start with random edges allocation, (ii) prioritize comparisons that connect disconnected components, (iii) compare dissimilar candidates to reduce uncertainty. Under some circumstances, it is enough to provide $\Theta(n*log(n))$ edges (see, for example, \cite{NegahbanOhShah2012RankCentrality}, \cite{Ailon2010ActiveLearningRanking}).

At the same time, we use a pointwise-based one as a baseline, calculating absolute score for each joke and deriving the winner by comparison of such scores between a pair of jokes.

\subsection{Global ranking}
Given a pair of candidates $(a,b)$ for the same constraint, our preference model predicts which candidate an audience would prefer. We aggregate outcomes for each pair into a global ranking using paired-comparison models (Bradley--Terry--Luce family) \cite{BradleyTerry1952RankAnalysisPairedComparisons} and Elo-style procedures using the Evalica library  \cite{Ustalov:25}.

\section{Experiments}
\label{sec:experiments}
\subsection{Within-dataset results}
\begin{table}[t]
\centering
\caption{10-fold CV accuracy for pointwise, NoBasis-ablation, and pairwise models.}
\label{tab:within}
\begin{tabular}{lccc}
\toprule
\textbf{Dataset} & \textbf{Pointwise} & \textbf{NoBasis} & \textbf{Pairwise} \\
\midrule
Reddit & 63\% & 41\% & 83\% \\
ScaleAI & 58\% & 32\% & 64\% \\
H.Arena & 45\% & 37\% & 77\% \\
\bottomrule
\end{tabular}
\end{table}

In Table \ref{tab:within}, we compare preference predictors trained on pairwise labels against baselines. \textbf{NoBasis} is a direct LLM-judge baseline: given a pair of jokes, an evaluator LLM is prompted only with the generic question ``which joke is better?'' (no feature basis); its predicted winner is compared to the human label to compute accuracy. Across datasets, our pairwise feature-based models are more reliable for predicting human choices. These results suggest that pairwise learning provides a stronger signal than absolute scoring for humor preferences, especially in the noisiest setting (H.Arena), where audience effects and style variance are high.

\subsection{Cross-dataset transfer}
\begin{table}[t]
\centering
\caption{Cross-dataset transfer accuracy for \textbf{pairwise} models (train on rows, test on columns). 50\% is random choice.}
\label{tab:transfer-pairwise}
\begin{tabular}{lccc}
\toprule
\textbf{Train $\backslash$ Test} & \textbf{Reddit} & \textbf{ScaleAI} & \textbf{H.Arena} \\
\midrule
Reddit  & 83\% & 52\% & 51\% \\
ScaleAI & 72\% & 64\% & 66\% \\
H.Arena  & 61\% & 61\% & 77\% \\
\bottomrule
\end{tabular}
\end{table}

\begin{table}[t]
\centering
\caption{Cross-dataset transfer accuracy for \textbf{pointwise} baselines (control). 50\% is random choice.}
\label{tab:transfer-pointwise}
\begin{tabular}{lccc}
\toprule
\textbf{Train $\backslash$ Test} & \textbf{Reddit} & \textbf{ScaleAI} & \textbf{H.Arena} \\
\midrule
Reddit  & 63\% & 51\% & 54\% \\
ScaleAI & 46\% & 58\% & 57\% \\
H.Arena  & 44\% & 52\% & 45\% \\
\bottomrule
\end{tabular}
\end{table}

We test whether a preference model trained on one dataset generalizes to another. Transfer is partial and asymmetric, indicating that humor preferences include both transferable components (e.g., punchline clarity, expectation violation) and community-specific biases. 

Overall, pairwise models show materially stronger transfer than pointwise baselines in most off-diagonal settings, suggesting that relative judgments capture more stable structure than absolute ratings.

\section{SemEval-2026 MWAHAHA Submission}
MWAHAHA evaluates systems using human preference battles in an arena setting. Our submission follows the same principle: generate a diverse candidate set, then select via a learned preference proxy for the listener.
We originally planned to use trial-phase feedback to calibrate the target audience profile for selection; however, the competition workflow did not enable using trial labels to adapt final-phase selection. We therefore treated H.Arena preferences as our best available proxy. For non-English subtasks, we use exactly the same workflow (including the same 17-feature humor basis), but requested LLMs to generate associations and candidate jokes directly in the target language.

Our system gained rank 1 at the English and Chinese subtasks of MWAHAHA with
1041 (95\% CI [1009, 1064]) and 1081 (95\% CI [1031, 1127]) ELO scores correspondingly, and rank 2 at the Spanish subtask with ELO score 1091 (95\% CI [1053, 1121]).

\section{Discussion}
Our findings support a preference-first view of humor generation: generating diverse candidates is increasingly easy for modern LLMs, while reliably selecting what a target audience prefers remains the key difficulty. This also explains why results in the literature can appear heterogeneous: evaluation depends strongly on rater pools, cultural context, and the measurement protocol. 
Empirically, LLMs can generate many grammatically valid ``joke-shaped'' candidates, but novelty and audience alignment are fragile. The tendency to repeat common templates is well documented in LLM joke generation experiments \cite{jentzsch23}. This supports investing compute into diverse generation and focusing research effort on preference modeling and selection.

An advantage of an interpretable feature basis is that it enables targeted error analysis: for example, we observe that certain features (e.g., Clear Punchline, Subverting Expectations) are consistently selected across datasets, while other features appear community-specific.

\section{Limitations and Ethics}
\paragraph{Subjectivity and demographic variation.}
Humor depends on culture, context, and the listener; thus, any single preference model risks overfitting to its annotator population. Humor can easily drift into toxic or offensive content. We recommend explicit safety filtering and constraint-based generation to keep outputs within task norms. A further limitation is evaluator circularity: our pipeline relies on LLMs for feature extraction/decomposition, which can introduce evaluator bias. However, we hope aligning the preference model with human choices should eliminate this effect.


\section{Conclusion}

We presented a system for SemEval-2026 Task~1 (MWAHAHA) that treats constrained humor generation as a two-stage pipeline: diverse candidate generation followed by pairwise preference-based selection. We release a new set of human pairwise humor judgments and propose an interpretable feature-based preference modeling approach. Across three datasets, pairwise preference models outperform pointwise baselines and transfer more robustly across domains. Overall, our results support the view that continued progress in humor generation will increasingly depend on improving evaluation and preference modeling, rather than solely improving raw text generation.


\bibliography{custom}
\appendix

\section*{Appendix A. Humor basis}
\label{sec:app_a}
\begin{lstlisting}[basicstyle=\ttfamily\fontsize{6}{6}\selectfont]
Clear Punchline: Ensure the joke delivers a strong, unmistakable punchline for maximum impact.
Wordplay with Purpose: Use puns or wordplay that serves the joke, rather than relying on repetition or forced cleverness.
Universality: Use references that are widely understood or relatable to broaden appeal.
Natural Dialogue: Employ conversational exchanges to make the joke feel organic and engaging.
Subtlety Over Obviousness: Favor subtle humor that allows audiences to connect the dots over jokes that spell everything out.
Avoid Cliche: Steer away from jokes that rely on overused wordplay or tired humor structures.
Fresh Perspective: Offer a novel or surprising angle on familiar situations to keep material original.
Exaggeration: Amplifying a characteristic, situation, or behavior to absurd levels to highlight its comedic potential.
Subverting Expectations: Twisting a familiar setup creates delight by catching the audience off guard.
Character-Driven Humor: Use established stereotypes or behaviors to anchor the joke and build richer scenarios.
Economy of Words: Be concise and efficient with language, trimming unnecessary details to maximize comedic payoff.
Self-Deprecation: Playfully targeting oneself can disarm the audience and make humor more relatable.
Satirical Edge: Employ satire to critique social trends or behaviors, adding depth to the humor.
Anthropomorphism: Attribute human qualities to non-human entities for humorous effect.
Clever Analogies: Use creative comparisons that link unrelated concepts for a surprising comedic twist.
Memorable Imagery: Create vivid or amusing mental pictures that stick with the audience.
Dark Humor: Making light of subjects that are generally considered serious, taboo, or morbid.
\end{lstlisting}

\section*{Appendix B. Pointwise feature vector example}
\label{sec:app_b}
\begin{lstlisting}[basicstyle=\ttfamily\fontsize{6}{6}\selectfont]
Joke: Putting air in your tires used to be free now its costs a dollar... Its called inflation.

Absolute scores:
Clear Punchline: 1.0
Wordplay with Purpose: 1.0
Universality: 1.0
Natural Dialogue: 0.0
Subtlety Over Obviousness: 0.6
Avoid Cliche: 0.6
Fresh Perspective: 0.0
Exaggeration: 0.0
Subverting Expectations: 0.0
Character-Driven Humor: 1.0
Economy of Words: 1.0
Self-Deprecation: 0.0
Satirical Edge: 0.2
Anthropomorphism: 0.0
Clever Analogies: 0.9
Memorable Imagery: 0.1
Dark Humor: 0.0
\end{lstlisting}

\section*{Appendix C. Pairwise feature vector example}
\label{sec:app_c}
\begin{lstlisting}[basicstyle=\ttfamily\fontsize{6}{6}\selectfont]
Joke A: "How many digits of pi do you know?" - "All of them... I just always forget the order!"
Joke B: A friend has a fear of pi. I keep telling him it's irrational, but he doesn't listen.

Relative scores (1.0 means A is closer, 0.0 means B is closer to match the rule)
Clear Punchline: 0.5
Wordplay with Purpose: 0.0
Universality: 0.05
Natural Dialogue: 1.0
Subtlety Over Obviousness: 0.45
Avoid Cliche: 0.55
Fresh Perspective: 0.95
Exaggeration: 0.75
Subverting Expectations: 0.95
Character-Driven Humor: 0.25
Economy of Words: 0.05
Self-Deprecation: 0.85
Satirical Edge: 0.35
Anthropomorphism: 0.25
Clever Analogies: 0.45
Memorable Imagery: 0.95
Dark Humor: 0.0
\end{lstlisting}

\end{document}